\documentclass{ecai} 

\usepackage{latexsym}
\usepackage{amssymb}
\usepackage{amsmath}
\usepackage{amsthm}
\usepackage{booktabs}
\usepackage{enumitem}
\usepackage{graphicx}
\usepackage{color}
\usepackage{bbm}
\usepackage{amssymb}
\usepackage{amsmath}
\usepackage{amsthm}
\usepackage{booktabs}
\usepackage{enumitem}
\usepackage{graphicx}
\usepackage{color}
\usepackage{url}
\usepackage{algorithm}
\usepackage{makecell}
\usepackage[noend]{algpseudocode}

\newtheorem{definition}{Definition}

\newcommand{\BibTeX}{B\kern-.05em{\sc i\kern-.025em b}\kern-.08em\TeX}

\begin{document}


\begin{frontmatter}


\paperid{123} 


\title{Towards Unsupervised Validation of Anomaly-Detection Models}


\author[A,B]{\fnms{Lihi}~\snm{Idan}\thanks{Corresponding Author. Email: lidan@hbs.edu, li49@tamu.edu}}

\address[A]{Harvard University}
\address[B]{Texas A\&M University}


\begin{abstract}
Unsupervised validation of anomaly-detection models is a highly challenging task. While the common practices for model validation involve a labeled validation set, such validation sets cannot be constructed when the underlying datasets are unlabeled. The lack of robust and efficient unsupervised model-validation techniques presents an acute challenge in the implementation of automated anomaly-detection pipelines, especially when there exists no prior knowledge of the model's performance on similar datasets. This work presents a new paradigm to automated validation of anomaly-detection models, inspired by real-world, collaborative decision-making mechanisms. We focus on two commonly-used, unsupervised model-validation tasks --- model selection and model evaluation --- and provide extensive experimental results that demonstrate the accuracy and robustness of our approach on both tasks.
\end{abstract}

\end{frontmatter}


\section{Introduction}
Anomaly detection (AD) is the task of identifying the minority of data
observations that deviate significantly from the majority of the observations.
While anomaly-detection tasks are challenging, they can be well-managed when the underlying datasets are labeled: training is performed using either supervised or semi-supervised models; the best model (\textit{i.e.} model selection) is chosen using a labeled validation set; and the performance of the model is assessed (\textit{i.e.} model evaluation) using a (second) labeled validation set.

The above no longer holds when the anomaly-detection task at hand is unsupervised and labeled datasets are not available. Though the pool of unsupervised anomaly-detection models has been steadily increasing over the past few years thus reducing the requirement of labeled training sets, two acute challenges remain: unsupervised model selection and unsupervised model evaluation. Due to the lack of a labeled validation set, standard model selection and evaluation practices cannot be applied while research on methods that do not require a labeled validation set is surprisingly scarce.

In this work, we aim to fill this gap by introducing a new approach to model selection and evaluation of anomaly-detection models which does not require any labeled data.   
Our approach is based on the following key ideas:

1. In cases where the ground truth is not available, a representative majority's opinion is a good-enough approximator for the ground truth.

2. One way to obtain a representative majority's opinion is building an \textit{Accurately-Diverse ensemble}: an ensemble of unsupervised models that sufficiently balances the ensemble's accuracy and diversity.

3. In order for an ensemble to be Accurately-Diverse, \textit{the ensemble’s decisions must exhibit both heterogeneity and homogeneity in a complementary manner}: we require a strong intra-ensemble agreement on the general trend of each ensemble member's predictions, while encouraging a strong intra-ensemble disagreement on the exact ordering of each ensemble member's predictions.


4. The exact ordering of a model's predictions is approximated by the ranked anomaly-score indices of the \textit{least distinctive observations in the dataset: non-extreme inliers.} The general trend of a model's predictions is approximated by its "fuzzy ranks": ranked anomaly-score clusters of the \textit{most distinctive observations in the dataset: strong outliers.} 
The level of agreement among ensemble members is approximated using a rank correlation metric, computed separately on fuzzy ranks of strong outliers and on exact ranks of non-extreme inliers. 

5. The rank correlation metric should be carefully designed so that it can both be applied to $M>2$ lists, and so that it can account for the unique structure of anomaly score lists compared to other ranked lists. We design multiple multi-way correlation metrics that are specifically suited to measuring the correlation among ranked anomaly scores.

We provide a thorough evaluation of our approach using ten publicly available datasets. Our experimental results demonstrate the following two claims:

1. An Accurately-Diverse ensemble yields better results than the average unsupervised anomaly-detection model; the results prove that using the Accurately-Diverse ensemble practically eliminates the need for a model-selection procedure of unsupervised anomaly-detection models. 

2. Our Accurately-Diverse-ensemble-based unsupervised evaluation metric yields results that are on par with those of supervised evaluation metrics; the results prove that the Accurately-Diverse ensemble can be used for unsupervised evaluation of anomaly detection models.

\textit{This work is the first to develop and experimentally test the ``complementary homogeneity-heterogeneity'' criteria: the criteria of a strong intra-ensemble agreement on the general trend of each ensemble member’s predictions and a strong intra-ensemble disagreement on the exact ordering of each ensemble member’s predictions, as a proxy for the ensemble's validity and its ability to be used for unsupervised, anomaly-detection model selection and evaluation.}
Importantly, from our experiments, it is clear that \textit{both} homogeneity \textit{and} heterogeneity on \textit{complementary} parts of the dataset are necessary for achieving a high degree of accuracy; ensembles that meet only one criterion, such as those described in prior work, demonstrate poor and unstable results.

\section{Related work}\label{sec:related}

As the limitations of unsupervised anomaly detection have been widely acknowledged, there have been multiple recent attempts to design more effective unsupervised AD models: probabilistic methods \cite{ecod,copola}, neural-network-based methods \cite{deepsv,gaal,alad} and even graph-based methods \cite{lunar}. Works such as \cite{eval1,eval2} provide benchmarks of the most prominent unsupervised anomaly-detection models as well as create practical rules of thumb on the best settings for using each model based on different criteria such as the dataset's domain and quality.

The use of ensembles for AD has been explored in multiple works. Yet, those works significantly differ from ours: either they assume a supervised or a semi-supervised setting where labels are available \cite{sup1,semisup1}; use ensembles based on feature diversification \cite{bagging1,bagging2}; use homogeneous ensembles \cite{xgboost,lightgbm,inne} or histogram-based ensembles \cite{loda}. Ensembles that do support fully unsupervised, heterogeneous ensemble members assume that the list of individual models \textit{is known in advance}; thus, they focus on methods for combining models' predictions \cite{combine,probs} or on researching optimal transformations that can be applied to individuals predictions \cite{suod,lscp} rather than on researching optimal ways for \textit{choosing} an optimal composition of the ensemble; that is, which models should be included in the ensemble. Importantly, no prior work aims at designing an AD ensemble model that can function as an \textit{unsupervised model selector}, while the latter is the main goal of Accurately Diverse ensembles.

The research on unsupervised evaluation methods of anomaly-detection models has been surprisingly scarce. The fully unsupervised evaluation approaches that we are aware of are the method described in \cite{ireosfull}, based on soft-margin classifiers, and the method described in \cite{emmv}, based on excess mass and mass volume curves. All other methods that we are aware of, such as the p-value-based evaluation method in \cite{pval}, require a labeled validation set.

Contrary to the above, our work focuses on a purely unsupervised setting in which ensemble members are heterogeneous and no feature transformation is performed.  
Our work is the first to develop and experimentally test the criteria of a strong
intra-ensemble agreement on the general trend of each ensemble member’s predictions and a strong intra-ensemble disagreement on the exact ordering of each ensemble
member’s predictions as a proxy for the ensemble's validity and its ability to both replace the model selection procedure of unsupervised anomaly-detection models and be utilized for unsupervised evaluation of anomaly-detection models.

\section{Accurately Diverse Ensembles}\label{sec:approx}

The diversity-accuracy tradeoff is an inherent challenge in ensemble-based approaches. In unsupervised settings, it becomes an even bigger challenge since the common practices for testing the accuracy of a model such as using a labeled validation set can not be applied. 
Our assumption is that when evaluating our ensemble, we do not have access to any source of "ground truth".
We, therefore, need to design a metric that will enable us to balance the accuracy and diversity of an ensemble without any access to labels. 

Our core idea used for designing such a metric is inspired by \cite{law}, which lists three requirements from a judicial ensemble: opinion heterogeneity, opinion homogeneity, and independence of errors. At first glance it is unclear how a single unsupervised ensemble can meet all three requirements: first, opinion homogeneity and heterogeneity, as well as opinion homogeneity and independence of errors, seem to be mutually exclusive. Second, in order to evaluate the independence-of-errors requirement we must have access to the ground truth of a subset of observations so we can determine which predictions were "mistakes". This requirement cannot be accommodated in the unsupervised case. Nevertheless, we claim that an unsupervised ensemble that meets all three requirements not only exists, but can also be easily identified using the following key observation:

\textit{To balance accuracy and diversity, the ensemble’s decisions must exhibit both heterogeneity and homogeneity in a complementary manner that highlights the general,
common shape of the ensemble member’s decision boundary
and at the same time blurs their individual peculiarities. 
This conceptual observation can be practically approximated by requiring a strong intra-ensemble agreement on the general trend of each member's predicted anomaly scores, while encouraging a strong intra-ensemble disagreement on the exact ordering of each member's predicted scores.
In such a case, the individual errors of the ensemble members will be sufficiently independent so that the aggregated decision coincides with the ground truth. We refer to such an ensemble as an Accurately-Diverse ensemble.}

\noindent
Specifically, for an ensemble to be Accurately-Diverse two conditions must hold:

(1) The ensemble members should strongly agree on \textit{high-level features} of \textit{highly-distinctive observations} in the dataset.

(2) The ensemble members should strongly disagree on \textit{low-level features} of \textit{lowly-distinctive observations} in the dataset.

Our main claims are the following:


\textbf{Claim 3.1.} In unsupervised AD settings where a supervised model-selection procedure --- a procedure that compares the performance of $N$ candidate AD models on a given dataset --- cannot be performed due to the lack of labeled data, an Accurately Diverse ensemble yields better results than the average anomaly-detection model thus eliminating the need for a model-selection procedure.

\textbf{Claim 3.2.} In unsupervised AD settings where the only model that can be used is a single model rather than an ensemble (for instance, due to regulatory requirements) yet due to the lack of a labeled validation set a supervised evaluation of the model cannot be performed, an Accurately-Diverse ensemble can be used to evaluate the model's predictions in an unsupervised manner, yielding results that are on par with those of supervised evaluation metrics.

Combining the two claims, once we have built an Accurately-Diverse ensemble we can use it in two ways: first, we can use an aggregation of the set of models that were selected for the ensemble as our unsupervised predictive model. 
Second, we can use the ensemble to evaluate other models in an unsupervised manner. In the next sections, we provide both the technical procedure for building an Accurately Diverse ensemble and algorithms for using the ensemble for the two above applications.  

\section{Building an Accurately Diverse Ensemble}
\subsection{Distinctive observations}


We define highly distinctive observations as follows:

\begin{definition}
A \textbf{highly distinctive observation} is an observation to which \textit{at least} one ensemble member gave a \textit{high} anomaly score. 
\end{definition}
Given the above definition, a natural definition of lowly-distinctive observations is the following:

A \textbf{lowly-distinctive observation} is an observation to which \textit{none} of the ensemble members gave a \textit{high} anomaly score.

Indeed, our first experiments were conducted under the above definition of lowly-distinctive observations. However, we found the results to be unsatisfactory. Upon further analysis, we have noticed that unsatisfactory results are obtained on datasets of a very specific type, which we refer to as "synthetic" anomaly-detection datasets: standard multi-way classification datasets where different classes are synthetically merged to create the outlier and inlier classes (\textit{e.g.} mnist \cite{mnist}, pendigits \cite{pendigits}, \textit{etc}). 

Our requirement of complimentary homogeneity and heterogeneity from the ensemble members is based on the key observation that 
the outlier class has well-defined characteristics that make its identification coincide with an "absolute truth" while the inlier class does not have such well-defined characteristics. A classic example is the task of financial transaction classification: it is rather easy to list common features of fraudulent transactions, while normal transactions lack such features and are rather described by negating fraudulent transactions' features. Thus, ranking inliers for their degree of normality is more of a subjective task than an objective one. This observation indeed holds in original anomaly-detection datasets (\textit{e.g.} fraud \cite{fraud}). However, in synthetic anomaly-detection datasets this observation no longer holds since the inlier class does have well-defined characteristics, independently of the outlier class and thus ranking of inliers becomes an objective task rather than a subjective one; for this reason, we cannot expect to see a strong intra-ensemble disagreement on the most extreme inliers since their identification in such datasets also constitutes an "absolute truth".

To accommodate that important observation, we change our definition of lowly-distinctive observations:

\begin{definition}
A \textbf{lowly-distinctive observation} is an observation to which \textit{non} of the ensemble members gave a \textit{high} anomaly score, but \textit{neither} member gave an \textit{extremely low} anomaly score. 
\end{definition}

\subsection{High-level features}\label{subsec:block}



Anomaly-detection models are usually just the first step in a pipeline; oftentimes, the anomaly scores predicted by the model will not serve as the final output of the pipeline but instead will serve as the input for another step in the pipeline, in which a human analyst assigns a given treatment to a subset of the observations in the dataset. 
In such cases, due to the large amount of time and cost of treating all the observations that received a high anomaly score by the model, the analyst will perform the treatment in a top-to-bottom manner. For instance, if the treatment is simply a manual validation of the top-ranked observations in terms of predicted anomaly scores, the analyst will first manually validate the top $\alpha n$-ranked observations on the list of predicted anomaly scores; then, depending on various factors such as time and degree of error tolerance, she will manually validate the top $2 \alpha n$-ranked anomalies; this process will continue until the top $\alpha=\eta \delta$-ranked observations are validated, where $\eta$ denotes the contamination factor, and for $1+\epsilon>=\delta>=1$. 


In such a setting, the main factor that determines the treatment probability of an observation is not its absolute position, but instead, the cluster --- a set of rank indices --- within which it lies on the ranked list. The most common formalization of clusters parametrizes $\alpha$ using $\eta$ and partitions the dataset into $\mathcal{C} = 4$ clusters:

\textbf{Cluster \#1:} $[1,\eta \gamma_1 n], 0<\gamma_1<1$: highest-confidence outliers.

\textbf{Cluster \#2:} $[(\eta \gamma_1 n)+1,\eta n]$: lowest-confidence outliers.

\textbf{Cluster \#3:} $[(\eta n)+1,\eta n \gamma_2], \gamma_2>1$: lowest-confidence inliers.

\textbf{Cluster \#4:} $[(\eta n \gamma_2)+1, n]$: highest-confidence inliers.

\noindent
In our experiments, we found $0.25<=\gamma_1<=0.5$,  $3<=\gamma_2<=5$ to work best. 

In a real-world production environment, the question of whether an observation, $i$, was mapped to rank cluster $1$ or rank cluster $\mathcal{C}$ is significantly more important than whether it was ranked in position $x$ or position $x+1$, since the decision whether $i$ will be treated before it is sent out to the next pipeline's node is solely determined by the cluster to which $i$ is mapped and the 
hyperparameters $\gamma_1,\gamma_2$. This illustrates the fact that oftentimes, the most informative features of a model's predictions are not \textit{low-level features} such as the exact scores each observation received by the model, but rather more \textit{high-level, generalizable features} such as the cluster to which the exact score was mapped. In the next subsection, we show how both low-level features such as rank index positions and high-level features such as rank cluster positions can be used to design new correlation metrics that better capture intra-ensemble agreement.





\subsection{Agreement among ensemble members}\label{subsec:corrs}

The simplest method to define agreement among ensemble members is via the intersection of their binary predictions.
Such a method, however, is too coarse-grained and thus might fail to capture the underlying structure of the decision-making mechanism of each ensemble member.
Thus, instead of using the binary prediction vectors of each model for measuring intra-ensemble agreement, we use the models' score vectors. We apply the rank transformation to the score vectors for normalization purposes so that the correlation is not biased toward one of the members. The task of measuring the agreement among ensemble models is therefore reduced to defining a proper notion of correlation between the $M$ ranked anomaly score lists, $\{r_m | m \in \mathcal{M}\}$.

\subsubsection{Rank correlation metrics}

The most commonly-used rank correlation metrics are Spearman's $\rho$ and Kendall's $\tau$. Kendall's $\tau$ measures correlation as the number of opposite ("discordant") pairs in the two lists. The notion of a "discordant" pair, expressed using $\xi$, is used to define Kendall's $\tau$ ($\tau_2^r$):
\begin{equation}
\label{eq:corrrank}
\fontsize{8}{8}\selectfont
\tau_2^r = \frac{\sum_{i=1}^{n} \sum_{j>i} \mathbbm{1}_{\xi^r_{r_1,r_2}(i,j)}}{n(n-1)/2}
\end{equation}
\begin{equation}
\label{eq:discrank}
\fontsize{8}{8}\selectfont
\xi_{r_1,r_2}^r(i,j) = \begin{cases}
       1 & \text{if } sgn(r_1[i]-r_1[j]) = sgn(r_2[i]-r_2[j]) \\
        0 & \text{if } sgn(r_1[i]-r_1[j]) = -sgn(r_2[i]-r_2[j])\\

        \end{cases} 
\end{equation}
For simplicity of notation, we denote observations $i,j$ using their index in the dataset, $D$. For instance, $r_1[i]$ denotes the ranked anomaly score which model $m_1$ predicted for the observation residing at the $i$th index of the dataset.

We argue that existing rank correlation methods cannot be used for accurately measuring the correlation between multiple lists that represent predicted, ranked anomaly scores. First, Kendall's $\tau$ implicitly assumes that all the observations have an equal contribution to the correlation between the two ranked lists. That implicit assumption oftentimes doesn't accurately represent the correlation we would like to capture between the predicted anomaly scores of two models. 
Assume that we have only two models in the ensemble and for two observations, $i$, $j$, we observe the following ranks:  $r_1[i]=1, r_1[j]=n-1,  r_2[i]=n-2, r_2[j]=2$. $i$ and $j$ will be considered a discordant pair, and their contribution to the final correlation will be 0. Now assume we are given the ranks of another pair, $i^*$, $j^*$: $r_1[i^*]=n/2, r_1[j^*]=n/2+1,  r_2[i^*]=n/2+1, r_2[j^*]=n/2$. $i^*$ and $j^*$ will also be considered a discordant pair, and their contribution to the correlation score will be the same as the contribution of $i$ and $j$. When attempting to quantify the correlation between two AD models using their anomaly scores, the discordance of $i$, $j$ should be penalized much heavier than the discordance of $i^*$,$j^*$. 

But there is another, more profound reason why existing rank correlation methods cannot fully capture the agreement among AD models. A discordant pair is defined to be one such as $sgn(f(r_1[i])-f(r_1[j])) = -sgn(f(r_2[i])-f(r_2[j]))$. In Kendall's $\tau$, $f$ is the identity function, $\mathcal{I}$. A discordant pair is a pair such that $i$ is ranked higher than $j$ in $r_1$, whereas $i$ is ranked lower than $j$ in $r_2$. We claim that such a definition of discordance is too fine-grained for measuring the agreement among anomaly-detection models.
Let us look at the following example:
\begin{equation}
\small
r_1[i]=\frac{\eta}{2}, ~~~ r_1[j]=\frac{\eta}{2}+1, ~~~ r_2[i]=2\eta+2, ~~~ r_2[j]=2\eta+1
\end{equation}
\begin{equation}
\small
r_1[i^*]=\frac{\eta}{2}, ~~~ r_1[j^*]=2\eta+1, ~~~ r_2[i^*]=2\eta+2, ~~~ r_2[j^*]=\frac{\eta}{3}
\end{equation}
\normalsize
For $i,j,i^*,j^* \in D$. Both $i,j$ and $i^*,j^*$ will be considered discordant pairs according to Kendall's $\tau$. But do those two notions of "oppositely ranked", the one represented by the relation of $i$ and $j$ and the one represented by the relation of $i^*$ and $j^*$, have a similar contribution to the agreement between the two models? we claim that they do not: the opposite rank relation of $i^*$ and $j^*$ in $r_1$ and $r_2$ is a much stronger indicator of a disagreement between the two models compared to the opposite rank relation of $i$ and $j$. 

Let us try to formalize this observation. In Subsection \ref{subsec:block}, we have discussed the rank cluster access pattern commonly used when analyzing AD models' output. 
Instead of determining the relation between a pair of observations according to the relation between their rank indices, we can determine the relation between them according to the rank clusters to which their rank indices are mapped. Defining the relation between observations using their clusters --- a generalization of their exact rank position on the ranked list, serves as the basis of a fuzzy rank correlation metric, a generalization of an exact rank correlation, in which for a pair of observations to negatively contribute to the correlation it must reflect an \textit{opposite rank cluster position relation, rather than an opposite rank index position relation.} A discordant pair in this case will defined as follows: $sgn(\beta(r_1[i])-\beta(r_1[j])) = -sgn(\beta(r_2[i])-\beta(r_2[j]))$ where $\beta$ is the cluster-mapping function $\beta: R -> C$, $|R|=n$ and $|C|=\mathcal{C}$.
That is, $f$ is now the cluster-mapping function instead of the identity function.
Assuming the existence of 4 clusters as described in Subsection \ref{subsec:block}, we can see that although according to the prior definition of discordance, both $i,j$ and $i^*,j^*$ are marked as discordant, according to the new discordance definition \textit{only} $i^*,j^*$ are marked as discordant, 
while $i,j$ are marked as concordant. 

\noindent
Combining the two observations, we present a generalization of Kendall's $\tau$, suited to measuring the correlation between two anomaly-detection models --- a weighted, fuzzy correlation metric based on rank clusters instead of rank indices:
\begin{equation}
\label{eq:corrcluster}
\scriptsize
\tau_2^c = \frac{\sum_{i=1}^{n} \sum\limits_{j>i} \Omega_{*}(w(\Omega(r_1[i],r_2[i])),w(\Omega(r_1[j],r_2[j])))\mathbbm{1}_{\xi_{r_1,r_2}^c(i,j)}}{\sum_{i=1}^{n} \sum_{j>i} \Omega_{*}(w(\Omega(r_1[i],r_2[i])),w(\Omega(r_1[j],r_2[j])))}
\end{equation}
\normalsize
\begin{equation}
\scriptsize
\label{eq:disccluster}
\xi_{r_1, r_2}^c(i,j)\!=\!\begin{cases}
       1&sgn(\beta(r_1[i])\!-\!\beta(r_1[j]))= sgn(\beta(r_2[i])\!-\!\beta(r_2[j])) \\
        0 &sgn(\beta(r_1[i])\!-\!\beta(r_1[j])= -sgn(\beta(r_2[i])\!-\!\beta(r_2[j]))\\

        \end{cases} 
\end{equation}
\normalsize
Each pair of observations is weighted: first, we compute an aggregated rank index of $i$ over models $m_1$,$m_2$ using an aggregation function $\Omega$ and then map the aggregated index into a weight, resulting in the term $\hat{\Omega}(i)=w(\Omega(r_1[i],r_2[i]))$. This term represents $i's$ aggregated distinctiveness score over all $M$ models. Next, we compute a similar aggregated weight for $j$. Finally, we aggregate both $i$'s and $j$'s weights using $\Omega_*$; this yields the final weight of the pair $i,j$. In our experiments, we set $\Omega_*$ to the $max()$ operator to bias the combined weight towards the more anomalous observation among $i$ and $j$. The reader is referred to Subsection \ref{subsec:put} and the Appendix \cite{appendix} for concrete realizations of $\Omega$ and $w$.

\subsubsection{Multi-way correlation metrics}

We now describe how to extend both $\tau_2^r$ and $\tau_2^c$ to the multi-way case that is needed for measuring the degree of correlation among the $M$ models in the ensemble.

Our \textbf{multi-way, exact correlation metric}, $\tau_{M^m}^r$, extends $\tau_2^r$ to the multi-way case using the notion of the "largest concordant set" where the agreement among the $M$ models is quantified as the largest subset of models that induces the same type of relation on $i,j$. As $\tau_2^c$, $\tau_{M^m}^r$ is weighted so the correlation can be biased towards certain observations based on their distinctiveness level (Subsection \ref{subsec:put}). A pseudocode of $\tau_{M^m}^r$ can be found in the Appendix \cite{appendix}.

Our \textbf{multi-way, fuzzy correlation metric}, $\tau_M^{c}$, extends $\tau_2^c$ by combining the advantages of exact-rank-based discordance and fuzzy-rank-based discordance thus balancing the correlation so it is neither too general nor too fine-grained using the following two key ideas:

1. Unlike $\xi_{r_1,r_2}^c$, we enable discordance that is based on opposite rank index positions; however, unlike $\xi_{r_1,r_2}^r$, discordance that is based on opposite rank index positions is only allowed between observations $i,j$ that belong to the \textit{same} cluster both in $r_1$ and $r_2$. That is, we only consider intra-cluster exact-rank-based discordance. Inter-cluster exact-rank-based discordance is not considered.

2. The definition of exact-rank-based discordance is relaxed; for each pair, $i,j$, the relaxation is proportional to the cluster size to which $i$ and $j$ are mapped.

The discordance level of $i$ and $j$ is measured as follows:
\begin{equation} 
\label{eq:disccomb}
\scriptsize
\xi_{r_1,r_2}^{rc}(i,j)\!=\!\begin{cases}
0 & \beta(r_1[i]) \neq \beta(r_2[i]) \lor \beta(r_1[j])\neq\beta(r_2[j])\\
1 & \beta(r_1[i])=\beta(r_2[i]) \land \beta(r_1[j])=\beta(r_2[j]) \\ & \land~\beta(r_1[i])\neq\beta(r_1[j])\\
       1 & \beta(r_1[i])=\beta(r_2[i]) =\beta(r_1[j])=\beta(r_2[j])~\land  \\ & 
       sgn(r_1[i]\!-\!r_1[j]) = sgn(r_2[i]\!-\!r_2[j]\pm \delta |\beta(r_1[i])|)
       \\
        0 & \beta(r_1[i])=\beta(r_2[i]) =\beta(r_1[j])=\beta(r_2[j])~\land 
        \\ &
      sgn(r_1[i]\!-\!r1[j])\!=\!-sgn(r_2[i]\!-\!r_2[j]\!\pm\!\delta |\beta(r_1[i])|)
       \\

        \end{cases} 
\end{equation}

\normalsize


Our multi-way, fuzzy correlation metric is based on the two-way discordance definition in $\xi_{r1,r2}^{rc}$, extended to the multi-way case using the notion of the "largest concordant set" as shown in Algorithm \ref{alg:rmc}:
the largest concordant set is the set of models $M' \subseteq \mathcal{M}$ such that for every two models in $M'$, $m^*,m^{**}$,  $\beta(r_{m^*}[i])=\beta(r_{m^{**}}[i])=c_1$ and  $\beta(r_{m^*}[j])=\beta(r_{m^{**}}[j])=c_2$, and either $c_1 \neq c_2$ or, $c_1 =c_2$ and within that cluster,  
       $sgn((r_{m^*}[i]-(r_{m^*}[j]) = sgn((r_{m^{**}}[i]-(r_{m^{**}}[j]\pm \delta |(c_1)|)$. Further details on both $\tau_M^{c}$ and $\xi_{r1,r2}^{rc}$ are given in the Appendix \cite{appendix}.

\begin{algorithm}[tb]
\caption{$\tau_M^{c}$}
 \noindent
 \small
\textbf{Input:} $\mathcal{M}$: $\{r_m | m \in \mathcal{M}\}$ $M$ ranked anomaly score lists\\
 \noindent
\textbf{Output:} $\tau_M^{c}$: a Fuzzy, multi-way correlation coefficient  \\
\begin{algorithmic}[1]
\For{\texttt{$i \in range(0,n)$}}
\State $ind_i=\Omega(i,\mathcal{M})$ 
\State $w_i=w(ind_i)$
\For{\texttt{$j \in range(i,n)$}}
\State $ind_j=\Omega(j,\mathcal{M})$ 
\State $w_j=w(ind_j)$
\State $w_{ij}=\Omega_*(w_i,w_j)$
\State $sum_w=sum_w+w_{ij}$
\State $smaller=[\mathcal{C}]$ 
\State $bigger=[\mathcal{C}]$
\State $equal=[\mathcal{C}][\mathcal{C}]$
\For{\texttt{$m \in range(0,\mathcal{M})$}}
\If{$(\lnot \beta(r_m[i])=\beta(r_m[j]))$}
\State {$equal[\beta(r_m[i])-1][\beta(r_m[j])-1] +=1$}
\Else 
\If{$|r_m[i]-r_m[j]|<=\delta *|\beta(r_m[i])|$}
\State {$smaller[\beta(r_m[i])-1] +=1$}
\State {$bigger[\beta(r_m[i])-1] +=1$}
\Else
\If{$r_m[i]<r_m[j]$}
\State {$smaller[\beta(r_m[i])-1] +=1$}
\Else
\State {$bigger[\beta(r_m[i])-1] +=1$}
\EndIf
\EndIf
\EndIf
\EndFor
\State {$max_{smaller}=\max(smaller)$}
\State {$max_{bigger}=\max(bigger)$}
\State {$max_{equal}=\max(equal)$}
\State {$max_{rel}=\max(max_{smaller}, max_{bigger}, max_{equal})$}
\State $corr=corr+(M-max_{rel})*w_{ij}$
\EndFor
\EndFor
\State $\tau_M^{c}=1-\frac{corr}{sum_w*(M-\lceil \frac{M}{\mathcal{C} +\mathcal{C}^2} \rceil)}$
\State \textbf{return} $\tau_M^{c}$
\end{algorithmic}
\label{alg:rmc}
\end{algorithm}


\subsection{Putting it all together}\label{subsec:put}

The key idea for measuring the extent to which an ensemble is Accurately Diverse is assessing the degree of both the exact rank correlation and fuzzy rank correlation of its members' predictions, each time using a different weighted version of the dataset. Specifically, we use two multi-way rank correlation metrics: an exact-rank correlation metric based on rank indices, $\tau_{M^m}^r$, and a fuzzy-rank correlation metric based on rank clusters, $\tau_M^{c}$. Each correlation metric is computed using a different set of weights, which bias the correlation towards a different type of observations in the dataset based on the observations' distinctiveness level: when computing a fuzzy-rank correlation metric based on rank clusters, we assign \textit{higher weights to observations which at least one model found to be highly anomalous}; on the other hand, when computing an exact-rank correlation metric based on rank indices we assign \textit{higher weights to observations which (a) all the models found to be inliers and (b) no model found to be an extreme inlier}.

The requirement for a strong intra-ensemble agreement on the general trend of each ensemble member's predictions is approximated by the ensemble obtaining a high degree of fuzzy rank correlation computed using a multi-model weighting scheme, $\hat{\Omega}^c$, that upweights strong outliers. The requirement for a strong intra-ensemble disagreement on the exact ordering of each ensemble member's predictions is approximated by the ensemble obtaining a low degree of exact rank correlation computed using a multi-model weighting scheme, $\hat{\Omega}^r$, that upweights non-extreme inliers.

As noted in Subsection \ref{subsec:corrs}, our multi-model weighting scheme, $\hat{\Omega}$, is composed of two components: a rank index aggregation function, $\Omega$, and a weighting scheme, $w$, applied to the aggregated rank index.



\textbf{Weighting scheme} The weighting scheme, $w$,  that we found to work best for upweighting highly-distinctive observations --- strong outliers, is an exponential weighting scheme:
\begin{equation}
w(i)  = 1- (1- (e^{-(\frac{i}{\delta \eta})^b}))
\end{equation}
where $\delta \in [1+\epsilon,2]$ and $b \in [2, 10]$. Such a weighting scheme assigns high weights only to strong, predicted outliers; the definition of "strong" can be controlled using $\delta$.

The weighting scheme that we found to work best for upweighting lowly distinctive observations is a bell-shaped weighting function based on the Subbotin distribution: a Gaussian-shaped curve with a plateau in its center. The size of the plateau can be set using $\lambda$ to control the amount of uniformity among highly-weighted observations:
\begin{equation}
w(i)  = e^{-(\frac{|i-n \mu|}{n \sigma})^\lambda}
\end{equation}
where $\mu \in [0.5, 0.75]$, $\sigma \in [0.1, 0.3]$, and $\lambda \in [2, 10]$. Such a weighting scheme assigns high weights only to "normal" inliers: inliers that are strong enough but are also not the most extreme inliers in the dataset.

\textbf{Rank index aggregation function} $\Omega$, the rank index aggregation function that we use must be such that is biased towards smaller rank indices; that is, biased towards ensemble members that assigned the observation the highest anomaly scores.  
The 
rank index aggregation 
function we found to yield the best results is the harmonic mean.

\vspace{5pt}
The multi-model weighting scheme that we found to work best for upweighting lowly distinctive observations is:
\begin{equation}
\hat{\Omega}^r(i)  =  e^{-(\frac{|\frac{M}{\sum{m=1}^{M} 1/{r_m[i]}}-n \mu|}{n \sigma})^\lambda}
\end{equation}
The multi-model weighting scheme that we found to work best for upweighting highly distinctive observations is:
\begin{equation}
\hat{\Omega}^c(i) = 1- (1- (e^{-(\frac{\frac{M}{\sum{m=1}^{M} 1/{r_m[i]}}
}{\delta \eta})^b}))
\end{equation}
To conclude, we assess the extent to which an ensemble is Accurately Diverse using two criteria: (1) a high degree of fuzzy rank correlation on strong outliers \textit{and} (2) a low degree of exact rank correlation on non-extreme inliers. The higher the correlation in (1) and the lower the correlation in (2),  the more Accurately Diverse the ensemble is.

\begin{algorithm}[tb]
\caption{$\mathcal{UED}$ score}
 \noindent
 \fontsize{8}{9}\selectfont
\textbf{Input:} $\mathcal{M}$: $M$ anomaly-detection  models, $D$: dataset\\
 \noindent
\textbf{Output:} $\mathcal{UED}$: evaluation score of model $m^*$ \\

\begin{algorithmic}[1]

\State {$E =BuildEnsemble(\mathcal{M})$}
\State {$r_{\mathcal{A}} =GetAggregatedRankedPredictions(E,D)$}
\State {$R(r_{\mathcal{A}}) =Rank(r_{\mathcal{A}})$}
\State {$r_{m^{*}} =RankPredict(m^*, D)$}
\For{\texttt{$i \in range(0,n)$}}
\State $d_i = \mathcal{D}(r_{m^*}[i], R(r_{\mathcal{A}}[i]))$
\State $c_i = \zeta(i, \mathcal{M})$
\State $w_i = w(\Omega(r_{m^*}[i], R(r_{\mathcal{A}}[i])))$
\State $score = score + (d_i*c_i*w_i)$
\EndFor
\State  $\mathcal{UED} = 1-(score / N)$
\State \textbf{return} $\mathcal{UED}$
\end{algorithmic}
\label{alg:alg1}
\end{algorithm}

\section{The "Unsupervised Ensemble Divergence" score}\label{sec:eval}
In Section \ref{sec:approx}, we developed the criteria that an ensemble of AD models must follow in order to be Accurately-Diverse and claimed that it will perform better than the average single model on unsupervised AD tasks. However, there could be situations where an ensemble model cannot be used. For instance, regulatory constraints might require the use of a single model. 
In such cases, though a model-selection procedure is not required, a model-evaluation method is crucial for assessing the true performance of the model. Because we assumed an unsupervised setting, supervised evaluation methods using a labeled validation set cannot be used. In this section, we use our Accurately-Diverse ensemble to design a new unsupervised evaluation metric for evaluating anomaly-detection models. 
\begin{figure}[tb]
\centering
\includegraphics[height=1in, width=3in]{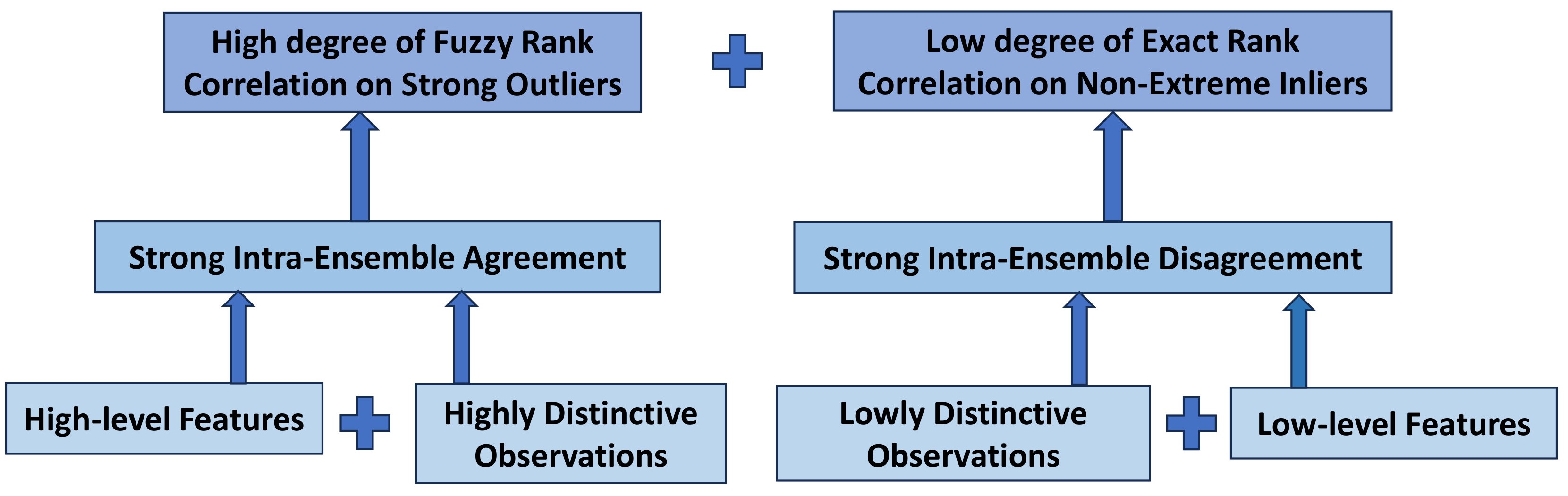}
\caption{Accurately Diverse Ensembles: System Diagram}
\label{fig:network}
\end{figure}
The core idea of the "Unsupervised Ensemble Divergence" score, $\mathcal{UED}$, is to use a distance metric, tailored specifically to AD tasks, measured between the prediction of the model to be evaluated, $m^*$, and the aggregated prediction of an Accurately-Diverse ensemble. Specifically, for each observation, $i$, we first compute $\mathcal{D}$, the distance between the Accurately-Diverse ensemble's aggregated prediction and the candidate model's prediction. We then compute $\zeta$, the ensemble's confidence level on observation $i$, approximated by the degree of unanimity of the ensemble members on the prediction of $i$. Finally, since we are evaluating an AD model and thus interested more in its accuracy on outliers, we compute the relative importance of $i$ to the score by assigning $i$ a weight that aggregates the position of $i$ in both the ranked score list predicted by the candidate model and the ranked score list predicted by the Accurately-Diverse ensemble. Here, $w$ is a rank-inverse weighting scheme in which if $r[i]>r[j]$, $w(i)<w(j)$. Concrete rank-inverse weighting schemes based on a cosine or a logarithmic reduction factor can be found in the Appendix \cite{appendix}.
The final contribution of $i$ to the overall evaluation score is proportional to a combination of the distance, the confidence, and the weight of observation $i$. Finally, the score is normalized using a distance-dependent normalization factor, $N$.

As seen in Algorithm \ref{alg:alg1}, we start by building an Accurately-Diverse ensemble as explained in previous sections and combining its members' predictions using an aggregation function (see Section \ref{sec:res}), resulting in the list of aggregated predictions, $r_{\mathcal{A}}$, for each observation in the dataset.  We then re-rank the aggregated-prediction list, resulting in a new ranked list, $R(r_{\mathcal{A}})$, and compute the distance between the following ranked lists: the ensemble's re-ranked aggregated predictions, $R(r_{\mathcal{A}})$, and the candidate model's ranked predictions, $r_{m^*}$. We have experimented with multiple rank distance metrics, and found a fuzzy rank distance metric based on rank clusters, $\mathcal{D}^c$, to yield the best results:
\begin{equation}
\mathcal{D}^c(r_{m^*},r_{\mathcal{A}}) = \sum_{i=1}^{n}|\beta(r_{m^*}[i]) - \beta(R(r_{\mathcal{A}}[i]))|
\end{equation}
After experimenting with multiple confidence metrics ($\zeta$) we have found the following metric to yield the best results:
\begin{equation}
\label{eq:median}
\zeta(i)= 1-\frac{\sum_{m \in \mathcal{M}}|\mathcal{MEDIAN}(\{r_{m}[i]\}_{m \in \mathcal{M}})-r_m[i]|)}{ (n-1)  \lfloor M/2 \rfloor }
\end{equation}
That is, the difference between each ensemble member's rank of observation $i$, and the ensemble's median rank of $i$.

The reader is referred to the Appendix \cite{appendix} for further details.
\begin{table*}[tb]

\centering

    \caption{A comparison of the Accurately-Diverse ensemble to the average single model (PR AUC, prec\textit{@n})}

        \setlength\tabcolsep{10pt}
                \setlength{\extrarowheight}{10pt}

       \begin{tabular}[tb]{|c|c|c|c|c|c|r|} 
    
    \hline
      \textbf{Dataset} &\shortstack{Best \\ensemble} & \shortstack{Worst \\ensemble}&  AS & RSPS&\shortstack{\% Improvement\\w.r.t RSPS}&\shortstack{\cite{loda},\cite{bagging2},\cite{suod},\cite{lscp} \\ (PR AUC)} \\ \hline

       smtp \cite{kdd} &.39, .55& 
        .02, .06& 
       .19, .35  & 
       .11,  .2 &
       254, 175&.2,.001,.09,.04
      \\   \hline
       mnist  \cite{mnist}   &.41,  .42&
     .15,  .17 & 
    .3,  .31& 
     .26,  .3 &
     57, 40&.19,.24,.34,.27
     \\ \hline
        
         backdoor  \cite{backdoor} &.48,  .45& 
        .04,  .06 & 
        .28,  .29&   
        .2,  .25 &
        140, 80&.02,.2,.13,.27
        \\ \hline
  
     gamma  \cite{gamma}   &.48,  .43&
     .35,  .38&
    .39,  .37& 
     .37,  .35  & 
     29, 23&.35,.33,.44,.43
     \\ \hline
     fraud  \cite{fraud}  &.29,  .34&
     .02,  .06 & 
    .12,  .19& 
     .12,  .18 &
     141, 88&.08,.001,.27,.15
     \\ \hline
      campaign  \cite{fraud} &.34,  .38&
     .13,  .13 & 
    .25,  .3& 
     .23,  .29 & 
     48, 31&.21,.15,.24,.22
     \\ \hline
      satellite  \cite{sat2}   &.6,  .55&
     .25,  .3 & 
    .49,  .46& 
     .44,  .44  & 
     36, 25&.5,.23,.45,.36
     \\ \hline
     pendigits  \cite{pendigits}   &.28,  .35&
     .03,  .03 & 
    .16,  .2& 
     .12,  .18  & 
     133, 94&.21,.05,.08,.08
     \\ \hline
      shuttle  \cite{sat2}  &.92,  .81&
     .13,  .15 & 
    .47,  .48& 
     .36,  .47  &
     155, 72&.1,.08,.51,.65
     \\ \hline

    \end{tabular} 
 
 \label{tab:res1}
\end{table*}

\section{Experimental results}\label{sec:res}


\noindent
Our first experimental task is to evaluate the performance of our Accurately-Diverse ensemble when used as a standalone unsupervised predictive model. This task is not straightforward, as it is not immediately clear which benchmarks are appropriate for the unsupervised setting. For instance, an inappropriate benchmarking methodology would be to compare the results of the Accurately-Diverse ensemble to those obtained by other unsupervised AD models and report the ensemble as having a high performance if its performance is better than the other models that we benchmark. This method is not a valid evaluation method as it is not representative of the true setting in which the ensemble will be used: specifically, when the analyst performs the model selection process in the unsupervised setting she has no way of knowing which model out of the $N$ candidate models performs the best. Thus, even if one of the $N$ models, $m'$, performs better than the Accurately-Diverse ensemble, this does not imply that the ensemble is inferior to $m'$ since $m'$ will probably not be selected as the model of choice. In fact, under our no-prior-knowledge assumption, it will be chosen only with a probability of $1/N$.
For the evaluation results to be representative of the true unsupervised setting in which our ensemble will be applied we compare our results to those of the average anomaly-detection model --- the \textit{average single model}. Assuming that we are given the option to choose a model out of $N$ candidate unsupervised AD models, the average single model can be evaluated in two ways:

1. Average Score (AS): evaluate the performance of each one of the $N$ candidate models using a supervised evaluation metric. Then average the results.

2. Randomly-Sampled Prediction Score (RSPS): given the predictions of each of the $N$ candidate models and an observation, $i$,  we randomly sample one of the $N$ predictions of $i$; by repeating this process for every observation, we form a new, randomly-sampled list of predictions. We then use a supervised metric to evaluate the performance of that list.

The idea behind AS and RSPS is simulating the real-world, unsupervised use case in which our ensemble will be used and in which, given $N$ candidate models, the analyst's choice of model is practically random.

Table \ref{tab:res1} compares our ensemble model results with the results of the average single model over different datasets.
For each dataset, given a pool of unsupervised models, we first build an Accurately-Diverse ensemble of size $M$, where, in our experiments,  $M=5$. We train the ensemble on the dataset and then use it to form predictions by aggregating the individual predictions of the ensemble members. After experimenting with multiple aggregation methods we found the arithmetic mean to perform best. We then re-rank the aggregated predictions and use the result as the final ensemble's output. Our pool of models is composed of $N\approxeq 25$ of the most commonly-used unsupervised AD models; 
specifically, we followed \cite{eval2} and used most of the models that they used, as well as some newer models
 \cite{gaal,lunar,alad}. All models were implemented using PyOD \cite{pyod}.
 The results of the average top Accurately-Diverse ensemble and the average bottom-Accurately-Diverse ensemble for each dataset are shown in Table \ref{tab:res1} and are compared against the results of the average single model approximated using the AS and RSPS. 
 For evaluation purposes, we use the PR AUC and prec\textit{@n} scores. 
The heuristic that we used in order to choose the top Accurately-Diverse ensembles is the following: we first sort the ensembles by the degree of their fuzzy rank correlation; then, out of the top-ranked ensembles in terms of fuzzy rank correlation, we choose those with the lowest exact rank correlation.

 As shown in Table \ref{tab:res1}, the top Accurately Diverse ensemble consistently outperforms the average single model using both the AS and RSPS. In addition, there is a significant difference in performance between the top Accurately-Diverse model and the bottom Accurately-Diverse model. 
 The results support Claim 3.1, according to which an Accurately-Diverse ensemble yields better results than the average anomaly-detection model. Thus, the process of creating an Accurately Diverse ensemble can be seen as the equivalent of a model-selection procedure in the unsupervised setting where no labeled data is available. 
Furthermore, even though existing AD ensemble models are not designed to function as model selectors per se, Table \ref{tab:res1} shows the PR AUC results of three state-of-the-art, fully-unsupervised AD ensemble models \cite{suod,lscp,loda} as well as a classic feature-bagging-based ensemble \cite{bagging2} on all datasets. Our results significantly outperform the results of all the ensemble methods used as baselines, demonstrating that our novel, "complementary homogeneity-heterogeneity"-based methodology is a superior methodology for ensemble building compared to prior methods.
\begin{table}[tb]

\centering

        \caption{Spearman correlation (PR AUC)}

        \setlength\tabcolsep{10pt}
                \setlength{\extrarowheight}{10pt}

       \begin{tabular}[b]{|c|c|c|c|c|c|c|c|c|c|c|r} 
    
    \hline
      \textbf{Dataset} &\textbf{Ours}& \cite{ireosfull}&\cite{emmv}   \\ \hline

      smtp \cite{kdd}&.9 &.72&.36  \\ \hline

         mnist  \cite{mnist}&.91&.75&.39 \\ \hline
         backdoor \cite{backdoor}&.88&.68& .32\\ \hline
        gamma  \cite{gamma}&.9&.61&.5  \\ \hline
                fraud  \cite{fraud}&.84& .7&.55 \\ \hline
                satellite  \cite{sat2,odd}&.88&.68& .34 \\ \hline
                campaign \cite{fraud} &.93&.83&.6  \\ \hline
                pendigits  \cite{pendigits}&.96&.88&.37  \\ \hline
                shuttle \cite{sat2} &.86&.74&.55  \\ \hline

    \end{tabular} 
   \label{tab:res2}

\end{table}

Table \ref{tab:res2} shows the Spearman correlation results between the $\mathcal{UED}$ score and the PR AUC score. For each dataset, we used the top-Accurately-Diverse ensembles and the procedure described in Algorithm \ref{alg:alg1} to evaluate the performance of each of the $N$ candidate models that \textit{were not selected to be part of the ensemble} thus forming a vector of unsupervised evaluation results. We also evaluated the performance of the $N$ models using the PR AUC, resulting in a vector of supervised evaluation results. We then computed the Spearman correlation between the two vectors. 
The multiplicative weighting scheme (Equation 4, Appendix \cite{appendix}) yielded the highest Spearman correlation results followed by the exponential weighting scheme (Equation 5, Appendix \cite{appendix}) with $\alpha=\delta \eta$ and $\delta \in [2, 4]$. 
The results demonstrate a very high correlation between the $\mathcal{UED}$ scores and the PR AUC scores when using a median-based confidence metric (Equation \ref{eq:median}). The results support Claim 3.2 according to which an Accurately-Diverse ensemble can be used to evaluate the results of other anomaly-detection models, yielding results that are on par with supervised evaluation metrics.
Finally, we compare our results to the results obtained by the methods in \cite{ireos,ireosfull,emmv}. As seen in Table \ref{tab:res2}, the results obtained using the method presented in \cite{ireos,ireosfull} are significantly lower than ours. The method presented in \cite{emmv} yielded extremely low results to the point of a non-existent correlation.

\section{Conclusion and broader impact}
Anomaly detection without access to labeled data is a highly challenging task. While the pool of unsupervised anomaly-detection models has been steadily increasing, as was evident while performing our experiments there exists no model that achieves high performance on \textit{all} the datasets. Analysts and practitioners thus face an acute problem: which model should be chosen out of all the available options? Moreover, how should the chosen model be evaluated so that the risk associated with the model's deployment can be correctly assessed? This work aims to provide a robust, generalizable, and above all --- accurate solution to the challenges of unsupervised model selection and evaluation for anomaly detection tasks. The novel idea of requiring the ensemble's decisions to exhibit both homogeneity \textit{and} heterogeneity in a complementary manner, an idea which we practically approximate by requiring a strong intra-ensemble agreement on the fuzzy anomalous ranks of strong outliers and a strong intra-ensemble disagreement on the
exact anomalous ranks of non-extreme inliers, is proven to serve as a reliable proxy for the ensemble's validity. 
We hope that the methodology presented in this work will not only provide a viable solution to the challenge of unsupervised model validation, but will also be used for addressing data-driven endeavors in other domains that can benefit from a new approach to balancing accuracy and diversity.

\newpage
\bibliography{ref}

\begin{thebibliography}{35}
\providecommand{\natexlab}[1]{#1}
\providecommand{\url}[1]{\texttt{#1}}
\expandafter\ifx\csname urlstyle\endcsname\relax
  \providecommand{\doi}[1]{doi: #1}\else
  \providecommand{\doi}{doi: \begingroup \urlstyle{rm}\Url}\fi

\bibitem[Aggarwal and Sathe(2015)]{combine}
C.~C. Aggarwal and S.~Sathe.
\newblock Theoretical foundations and algorithms for outlier ensembles.
\newblock \emph{SIGKDD explorations newsletter}, 17\penalty0 (1):\penalty0 24--47, 2015.

\bibitem[Alimo{\u{g}}lu and Alpaydin(2001)]{pendigits}
F.~Alimo{\u{g}}lu and E.~Alpaydin.
\newblock Combining multiple representations for pen-based handwritten digit recognition.
\newblock \emph{Turkish Journal of Electrical Engineering and Computer Sciences}, 9\penalty0 (1):\penalty0 1--12, 2001.

\bibitem[Bandaragoda et~al.(2018)]{inne}
T.~R. Bandaragoda et~al.
\newblock Isolation-based anomaly detection using nearest-neighbor ensembles.
\newblock \emph{Computational Intelligence}, 34\penalty0 (4):\penalty0 968--998, 2018.

\bibitem[Bates et~al.(2023)Bates, Cand{\`e}s, Lei, Romano, and Sesia]{pval}
S.~Bates, E.~Cand{\`e}s, L.~Lei, Y.~Romano, and M.~Sesia.
\newblock Testing for outliers with conformal p-values.
\newblock \emph{The Annals of Statistics}, 51\penalty0 (1):\penalty0 149--178, 2023.

\bibitem[Chen and Guestrin(2016)]{xgboost}
T.~Chen and C.~Guestrin.
\newblock Xgboost: A scalable tree boosting system.
\newblock In \emph{International Conference on Knowledge Discovery and Data Mining}, pages 785--794, 2016.

\bibitem[Edlin(2020)]{law}
D.~E. Edlin.
\newblock \emph{Common Law Judging: Subjectivity, impartiality, and the making of Law}.
\newblock University of Michigan Press, 2020.

\bibitem[Emmott et~al.(2015)]{gamma}
A.~Emmott et~al.
\newblock A meta-analysis of the anomaly detection problem.
\newblock \emph{arXiv preprint arXiv:1503.01158}, 2015.

\bibitem[Gao and Tan(2006)]{probs}
J.~Gao and P.-N. Tan.
\newblock Converting output scores from outlier detection algorithms into probability estimates.
\newblock In \emph{ICDM}, 2006.

\bibitem[Goix(2016)]{emmv}
N.~Goix.
\newblock How to evaluate the quality of unsupervised anomaly detection algorithms?
\newblock \emph{arXiv preprint arXiv:1607.01152}, 2016.

\bibitem[Goldstein and Uchida(2016)]{eval1}
M.~Goldstein and S.~Uchida.
\newblock A comparative evaluation of unsupervised anomaly detection algorithms for multivariate data.
\newblock \emph{PloS one}, 11\penalty0 (4):\penalty0 e0152173, 2016.

\bibitem[Goodge et~al.(2022)Goodge, Hooi, Ng, and Ng]{lunar}
A.~Goodge, B.~Hooi, S.-K. Ng, and W.~S. Ng.
\newblock Lunar: Unifying local outlier detection methods via graph neural networks.
\newblock In \emph{AAAI}, 2022.

\bibitem[Han et~al.(2022)]{eval2}
S.~Han et~al.
\newblock Adbench: Anomaly detection benchmark.
\newblock In \emph{Advances in Neural Information Processing Systems}, pages 32142--32159, 2022.

\bibitem[Idan(2024)]{appendix}
L.~Idan.
\newblock Appendix: Towards unsupervised model validation.
\newblock osf.io/naq94, 2024.

\bibitem[Ke and thers(2017)]{lightgbm}
G.~Ke and thers.
\newblock Lightgbm: A highly efficient gradient boosting decision tree.
\newblock In \emph{NeurIPS}, 2017.

\bibitem[Lazarevic and Kumar(2005)]{bagging1}
A.~Lazarevic and V.~Kumar.
\newblock Feature bagging for outlier detection.
\newblock In \emph{International Conference on Knowledge Discovery in Data Mining}, 2005.

\bibitem[LeCun et~al.(1998)LeCun, Bottou, Bengio, and Haffner]{mnist}
Y.~LeCun, L.~Bottou, Y.~Bengio, and P.~Haffner.
\newblock Gradient-based learning applied to document recognition.
\newblock \emph{Proceedings of the IEEE}, 86\penalty0 (11):\penalty0 2278--2324, 1998.

\bibitem[Li et~al.(2020)]{copola}
Z.~Li et~al.
\newblock Copod: copula-based outlier detection.
\newblock In \emph{ICDM}, 2020.

\bibitem[Li et~al.(2022)]{ecod}
Z.~Li et~al.
\newblock Ecod: Unsupervised outlier detection using empirical cumulative distribution functions.
\newblock \emph{IEEE TKDE}, 2022.

\bibitem[Liu et~al.(2019)Liu, Li, Zhou, Jiang, Sun, Wang, and He]{gaal}
Y.~Liu, Z.~Li, C.~Zhou, Y.~Jiang, J.~Sun, M.~Wang, and X.~He.
\newblock Generative adversarial active learning for unsupervised outlier detection.
\newblock \emph{IEEE TKDE}, 32\penalty0 (8):\penalty0 1517--1528, 2019.

\bibitem[Marques et~al.(2015)]{ireos}
H.~Marques et~al.
\newblock On the internal evaluation of unsupervised outlier detection.
\newblock In \emph{International Conference on scientific and statistical database management}, 2015.

\bibitem[Marques et~al.(2020)]{ireosfull}
H.~Marques et~al.
\newblock Internal evaluation of unsupervised outlier detection.
\newblock \emph{ACM Transactions on Knowledge Discovery from Data}, 14\penalty0 (4):\penalty0 1--42, 2020.

\bibitem[Moustafa and Slay(2015)]{backdoor}
N.~Moustafa and J.~Slay.
\newblock A comprehensive data set for network intrusion detection systems.
\newblock In \emph{Military Communications and Information Systems Conference}, 2015.

\bibitem[Nguyen et~al.(2010)]{bagging2}
H.~V. Nguyen et~al.
\newblock Mining outliers with ensemble of heterogeneous detectors on random subspaces.
\newblock In \emph{Database Systems for Advanced Applications}, 2010.

\bibitem[Pang et~al.(2019)Pang, Shen, and van~den Hengel]{fraud}
G.~Pang, C.~Shen, and A.~van~den Hengel.
\newblock Deep anomaly detection with deviation networks.
\newblock In \emph{KDD}, 2019.

\bibitem[Pevn{\`y}(2016)]{loda}
T.~Pevn{\`y}.
\newblock Loda: Lightweight on-line detector of anomalies.
\newblock \emph{Machine Learning}, 102:\penalty0 275--304, 2016.

\bibitem[Rayana(2016)]{odd}
S.~Rayana.
\newblock Odds library.
\newblock \emph{Stony Brook University, Department of Computer Sciences}, 2016.

\bibitem[Ruff et~al.(2018)Ruff, Vandermeulen, Goernitz, Deecke, Siddiqui, Binder, M{\"u}ller, and Kloft]{deepsv}
L.~Ruff, R.~Vandermeulen, N.~Goernitz, L.~Deecke, S.~A. Siddiqui, A.~Binder, E.~M{\"u}ller, and M.~Kloft.
\newblock Deep one-class classification.
\newblock In \emph{ICML}, 2018.

\bibitem[Srinivasan(1993)]{sat2}
A.~Srinivasan.
\newblock {Landsat Satellite}.
\newblock UCI Machine Learning Repository, 1993.

\bibitem[Stolfo et~al.(1999)]{kdd}
S.~Stolfo et~al.
\newblock {KDD Cup 1999 Data}.
\newblock UCI Machine Learning Repository, 1999.

\bibitem[Vargaftik et~al.(2021)Vargaftik, Keslassy, Orda, and Ben-Itzhak]{sup1}
S.~Vargaftik, I.~Keslassy, A.~Orda, and Y.~Ben-Itzhak.
\newblock Rade: resource-efficient supervised anomaly detection using decision tree-based ensemble methods.
\newblock \emph{Machine Learning}, 110\penalty0 (10):\penalty0 2835--2866, 2021.

\bibitem[Zenati et~al.(2018)Zenati, Romain, Foo, Lecouat, and Chandrasekhar]{alad}
H.~Zenati, M.~Romain, C.-S. Foo, B.~Lecouat, and V.~Chandrasekhar.
\newblock Adversarially learned anomaly detection.
\newblock In \emph{ICDM}, 2018.

\bibitem[Zhao and Hryniewicki(2018)]{semisup1}
Y.~Zhao and M.~K. Hryniewicki.
\newblock Xgbod: improving supervised outlier detection with unsupervised representation learning.
\newblock In \emph{IJCNN}, 2018.

\bibitem[Zhao et~al.(2019{\natexlab{a}})Zhao, Nasrullah, and Li]{pyod}
Y.~Zhao, Z.~Nasrullah, and Z.~Li.
\newblock Pyod: A python toolbox for scalable outlier detection.
\newblock \emph{Journal of Machine Learning Research}, 20\penalty0 (96):\penalty0 1--7, 2019{\natexlab{a}}.

\bibitem[Zhao et~al.(2019{\natexlab{b}})]{lscp}
Y.~Zhao et~al.
\newblock Lscp: Locally selective combination in parallel outlier ensembles.
\newblock In \emph{SDM}, 2019{\natexlab{b}}.

\bibitem[Zhao et~al.(2021)]{suod}
Y.~Zhao et~al.
\newblock Suod: Accelerating large-scale unsupervised heterogeneous outlier detection.
\newblock \emph{Proceedings of Machine Learning and Systems}, 3:\penalty0 463--478, 2021.

\end{thebibliography}

\end{document}